\documentclass[cameraready]{Interspeech}
\title{findsylls: A Language-Agnostic Toolkit for Syllable-Level Speech Tokenization and Embedding}

\author[affiliation={1}, orcid=0009-0004-5200-553X]{H\'ector Javier}{V\'azquez Mart\'inez}
\address{
    $^1$ Department of Linguistics, University of Pennsylvania, USA %\\
}

\email{hjvm@upenn.edu}%, second@companyA.com, third@companyB.ai}

\keywords{syllable segmentation, speech tokenization, speech representations, benchmarking, cross-lingual speech processing}

\usepackage{comment}
\usepackage{multirow}

\begin{document}

\maketitle

% the abstract here must exactly match the abstract entered into the paper submission system
\begin{abstract}
Syllable-level units offer compact and linguistically meaningful representations for spoken language modeling and unsupervised word discovery, but research on syllabification remains fragmented across disparate implementations, datasets, and evaluation protocols. We introduce findsylls, a modular, language-agnostic toolkit that unifies classical syllable detectors and end-to-end syllabifiers under a common interface for syllable segmentation, embedding extraction, and multi-granular evaluation. The toolkit implements and standardizes widely used methods (e.g., Sylber, VG-HuBERT) and allows their components to be recombined, enabling controlled comparisons of representations, algorithms, and token rates. We demonstrate findsylls on English and Spanish corpora and on new hand-annotated data from Kono, an underdocumented Central Mande language, illustrating how a single framework can support reproducible syllable-level experiments across both high-resource and under-resourced settings.
\end{abstract}

\section{Introduction}

Syllables provide a natural temporal unit for modeling speech, sitting between short analysis frames and longer lexical units such as words. Syllable-level representations have recently been shown to support efficient spoken language modeling, self-supervised word discovery, and downstream speech tasks, by drastically reducing sequence length while preserving linguistic structure \cite{choSylberSyllabicEmbedding2024, baadeSyllableLMLearningCoarse2024, wangUnsupervisedSpeechRecognition2025,leeScalingSpokenLanguage2026}. %At the same time, the practical ecosystem for syllable-level processing remains fragmented: Classical convex hull-based methods, oscillator models, sonority-based approaches, and self-supervised syllable tokenizers are implemented in disparate codebases, evaluated on different corpora, and rarely compared side-by-side beyond a handful of high-resource languages.

Self-supervised learning (SSL) frameworks such as VG-HuBERT \cite{pengWordDiscoveryVisually2022, pengSyllableDiscoveryCrossLingual2023}, SD-HuBERT \cite{choSDHuBERTSentenceLevelSelfDistillation2024}, Sylber \cite{choSylberSyllabicEmbedding2024} and SyllableLM \cite{baadeSyllableLMLearningCoarse2024} demonstrate that syllabic structure can be induced from large quantities of unlabeled audio, yielding syllable-granular tokens at a rate of 4-5Hz \cite{libermanSyllableRhythmEnglish2023} and support competitive performance in spoken language modeling.  Recent work on syllabic spoken language models further shows that syllable-level tokenization can match or surpass high-frame-rate SSL tokens on spoken language understanding benchmarks while reducing training time more than twofold and FLOPs by approximately fivefold, underscoring the practical scalability benefits of syllabic units for long-context modeling \cite{leeScalingSpokenLanguage2026}.  However, these models inherit the computational cost and pretraining biases of their base encoders, and their “syllable” units emerge only after multi-stage distillation from frame-level features.  Recent work in unsupervised lexicon learning further shows that mismatches between a SSL model’s pretraining languages and the target language can reduce segmentation accuracy by 20–30\%, and that representation quality, rather than clustering algorithms, is the dominant bottleneck even for English and Mandarin \cite{adendorffUnsupervisedLexiconLearning2025}.  %This dependence on large, language-specific pretraining challenges the ideal of “zero-resource” speech processing and raises questions about generalization to truly unseen or typologically distant languages.

%In parallel, work on unsupervised lexicon learning has highlighted that representation quality and pretraining conditions are critical bottlenecks for zero-resource speech processing \cite{adendorffUnsupervisedLexiconLearning2025}. What is currently missing is not yet another syllable model, but a shared infrastructure for implementing, recombining, and evaluating existing approaches in a comparable way across languages, including under-documented ones.

Several decades of research on amplitude envelopes \cite{mermelsteinAutomaticSegmentationSpeech1975, xieRobustAcousticbasedSyllable2006}, sub-band correlation \cite{wangRobustSpeechRate2007}, sonority profiles \cite{yuanRobustSpeakingRate2010, ludusanAutomaticSyllableSegmentation2016}, and rhythm-guided oscillators \cite{zhangSpeechRhythmGuided2009} have produced broadly cross-lingual methods for syllable segmentation that operate directly on the signal, often without large-scale supervised training. These methods are attractive for low-resource and cross-lingual settings, yet they are typically distributed as isolated scripts or prototypes, often tuned to specific corpora and difficult to integrate into modern speech processing pipelines. %Likewise, recent SSL-based syllabifiers are released with bespoke interfaces and evaluation setups. As a result, there is currently no standard toolkit that (i) unifies classical envelope-based algorithms and modern representation-driven syllabifiers, (ii) exposes a common interface for syllable segmentation, embedding, and evaluation, and (iii) facilitates systematic comparison across languages, speaking styles, and downstream tasks, from widely studied languages to under-documented ones such as Kono
%As a result, there is currently no standard toolkit that (i) unifies classical envelope-based algorithms and modern representation-driven syllabifiers, (ii) exposes a common interface for syllable segmentation, embedding, and evaluation, and (iii) facilitates systematic comparison across languages, speaking styles, and downstream tasks.
As a result, it remains difficult to (i) reproduce prior syllabification results, (ii) compare methods under matched datasets/metrics, and (iii) run controlled ablations that separate representation quality from segmentation choices.

In this paper, we address this gap by introducing \texttt{findsylls}\footnote{\url{https://github.com/hjvm/findsylls} \\ \url{https://pypi.org/project/findsylls/}}, a modular, language-agnostic toolkit for speech tokenization, embedding, and processing through syllabic units. The toolkit organizes envelope computation, feature extraction, and segmentation algorithms into interoperable modules, enabling mix-and-match experimentation and reuse across languages, corpora, and downstream tasks. We benchmark classical and representation-driven syllabifiers on English and Spanish adult and child-directed speech, as well as on new hand-annotated data from Kono, an under-documented Central Mande language, comparing segmentation accuracy, token rates, and computational cost. These experiments illustrate how a single framework can support reproducible syllable-level research in both high-resource and under-resourced settings.

Our contributions are as follows:
\begin{itemize}
    \item We introduce \texttt{findsylls}, an open-source toolkit that unifies classical amplitude-envelope syllable segmentation algorithms and recent neural syllabifiers under a single interface for segmentation, syllable-level embedding, and evaluation.
    \item We provide a reusable evaluation pipeline for syllable nucleus, boundary, and span detection against time-aligned annotations, including corpus-level aggregation tools suitable for large-scale benchmarking.
    \item We conduct a systematic comparison of envelope- and SSL-based syllabifiers on English, Spanish, and on newly annotated Kono data, analyzing segmentation accuracy, token rate, and computational cost.
    \item We position \texttt{findsylls} as shared infrastructure for syllable-level speech processing, designed to make it easy for the community to implement, recombine, and evaluate existing and future syllabification methods on both well-studied and under-documented languages.
\end{itemize}

\section{Methods}

\texttt{findsylls} organizes syllable processing into three interoperable modules: (i) envelope computation, (ii) frame-level feature extraction, and (iii) segmentation algorithms. Together, these modules cover both classical signal-processing approaches and recent representation-driven methods, while allowing their components to be recombined within a single framework. %This section summarizes the algorithms currently implemented.
\subsection{Envelope computation: classical syllable cues}

The first module in \texttt{findsylls} computes amplitude envelopes that capture syllabic rhythm using classical signal-processing techniques.  The toolkit provides several envelope variants that implement and extend long-standing approaches to syllable detection: RMS energy, low-pass filtered energy, Hilbert-transform envelopes, spectral band subtraction (SBS) envelopes \cite{libermanSyllables2022}, as well as a neurophysiologically inspired theta (oscillator) envelope \cite{rasanenPrelinguisticSegmentationSpeech2018}, with more methods to come.  These methods all produce a smoothed energy contour over time from which syllable nuclei can be detected as local maxima, following the convex-hull \cite{mermelsteinAutomaticSegmentationSpeech1975} and peak selection tradition.

%\texttt{findsylls} also includes a theta-oscillator envelope inspired by neurophysiological accounts of cortical entrainment to speech in the 3–8\,Hz band \cite{rasanenPrelinguisticSegmentationSpeech2018}.  This method decomposes the signal into gammatone sub-bands and drives damped oscillators tuned to syllabic frequencies, then combines their outputs into an envelope whose peaks tend to align with syllable nuclei, as in prior oscillator-based models.  %Across all envelope types, peak detection uses an automatically calibrated look-ahead based on a minimum syllable duration rather than fixed ad-hoc parameters, improving robustness across datasets and speaking rates.

\subsection{Feature extraction: self-supervised and classical representations}

The second module focuses on frame-level feature extraction, decoupled from any particular segmentation or frame-grouping strategy.  Classical extractors include MFCC and log mel-spectrograms at typical frame rates (e.g., 20ms), providing simpler acoustic representations that have historically been used in speech and phonetic research.  These features can be used on their own or as inputs to feature-based segmentation algorithms such as MinCut \cite{pengSyllableDiscoveryCrossLingual2023,baadeSyllableLMLearningCoarse2024} or greedy cosine merging \cite{choSylberSyllabicEmbedding2024}.

Complementing these, \texttt{findsylls} supports several self-supervised encoders whose internal representations have been shown to encode syllabic structure: HuBERT \cite{hsuHuBERTSelfSupervisedSpeech2021} trained on large-scale English audio, VG-HuBERT \cite{pengWordDiscoveryVisually2022,pengSyllableDiscoveryCrossLingual2023} further fine-tuned on audio–image pairs, and Sylber \cite{choSylberSyllabicEmbedding2024} distilled explicitly to produce syllabic embeddings.  Each extractor returns high-dimensional feature matrices, without imposing any particular segmentation.  The feature-based segmentation methods published along with these models are implemented in the third module and can be used interchangeably with any of the feature extraction methods implemented in \texttt{findsylls}.

By making classical and self-supervised features available through a common interface, the toolkit allows researchers to disentangle how much of syllabification performance arises from the representation source versus the segmentation algorithm applied on top.

\subsection{Segmentation algorithms and modular recombination}
\label{subsec:segmenters-and-modularity}

The third module comprises segmentation algorithms that operate on either envelopes or feature sequences and output syllable intervals. For envelope-based pipelines, \texttt{findsylls} implements Billauer's \texttt{peakdetect} algorithm \cite{billauerPeakdetPeakDetection2009} over the computed amplitude envelopes, yielding onset, nucleus, and offset times per syllable. For feature-based pipelines, each method reads a frame-level cue off the representation and applies a dedicated segmenter: a greedy merge over cosine similarity to a running segment prototype \cite{choSylberSyllabicEmbedding2024}, optimized MinCut over a feature self-similarity matrix (featSSM) \cite{baadeSyllableLMLearningCoarse2024,pengSyllableDiscoveryCrossLingual2023}, and thresholding of \texttt{[CLS]}-token attention \cite{pengWordDiscoveryVisually2022,choSDHuBERTSentenceLevelSelfDistillation2024}.

To support interoperability with envelope-based methods, \texttt{findsylls} exposes each cue as a pseudo-envelope, a scalar time series, decoupling it from its native segmenter. The cosine cue is each frame's similarity to a local prototype of its neighboring frames; featSSM is each frame's global mean similarity to all other frames; the \texttt{[CLS]} cue thresholds and reduces multi-head attention to one value per frame. Each trace is normalized and passed to \texttt{peakdetect} in place of a classical amplitude envelope, so the same peak-picking logic operates across classical and neural cues within a unified interface.
  
\section{The \texttt{findsylls} toolkit}

A key design goal of \texttt{findsylls} is that these algorithms are not tied to any specific feature extractor.  The greedy cosine strategy originally used in Sylber can be applied to HuBERT or VG-HuBERT features, enabling direct tests of how “Sylber-style” segmentation behaves with alternative encoders.  Similarly, the MinCut algorithm optimized with the VG-HuBERT and SyllableLM work can be driven by self-similarity matrices from VG-HuBERT, Sylber, HuBERT, or even classical acoustic features.  This modularity permits controlled ablations that swap only the representation or only the segmentation algorithm, clarifying their respective contributions to syllable boundary quality and token rates.

Beyond boundary detection, \texttt{findsylls} offers a syllable-level embedding pipeline that combines any segmenter with any feature extractor and aggregates features within each syllable using mean, max, median, or onset–nucleus–coda pooling, producing syllabic tokens suitable for downstream modeling.  For evaluation, \texttt{findsylls} provides a common interface for computing nucleus, boundary, and span F1 against time-aligned TextGrid annotations, making it straightforward to benchmark any configuration's performance on new corpora or languages as reference annotations become available.
%A shared evaluation framework then quantifies nucleus, boundary, and span F1 scores against annotated TextGrids across multiple corpora and languages, providing a common testbed for both replicated and newly recombined syllabification systems.

To demonstrate the utility of \texttt{findsylls}, we benchmark classical and neural syllabifiers on a diverse set of corpora spanning English, Spanish, and Kono (Table~\ref{tab:data}).

\begin{table}[h]
\footnotesize
\centering
\caption{Evaluation corpora.}
\label{tab:data}
\begin{tabular}{llcc}
\toprule
Name & Type & Hours & \# Syllables \\
\midrule
LS-100h & Adult read (English) & 100 & 1374476 \\
WikiSpanish & Adult read (Spanish) & 25 & 530816 \\
TIMIT & Adult read (English) & 5 & 63265 \\
PHC & Child-directed (English) & 0.84 & 10631 \\
Ornat-Swingley & Child-directed (Spanish) & 0.24 & 2755 \\
Brent & Child-directed (English) & 0.11 & 1487 \\
Kono & Fieldwork & 0.07 & 636 \\
\bottomrule
\end{tabular}
\end{table}

\begin{table*}[t]
\centering
\small
\setlength{\tabcolsep}{4pt}
\caption{
Segmentation performance averaged over seven corpora 
(Brent, Kono, LibriSpeech, Ornat, PhiladelphiaHomeCorpus, TIMIT, WikiSpanish).
Metrics are syllable-weighted Precision (Prec), Recall (Rec), and F1.
Nucleus = syllable peak detection; 
Boundary = inter-syllabic boundary detection; 
Span = full syllable interval recovery.
$\uparrow$ indicates higher is better; $\downarrow$ lower is better.
Tok/s = predicted syllable nuclei per second of audio (lower = more compression).
RTFx = inverse real-time factor (total\_audio\_duration / elapsed\_time) reported to illustrate relative throughput under a fixed hardware/software setup.
}
\label{tab:full_method_comparison}
\begin{tabular}{lll|ccc|ccc|ccc|cc}
\toprule
\multirow{2}{*}{Features} & \multirow{2}{*}{Envelope} & \multirow{2}{*}{Segmentation} 
& \multicolumn{3}{c|}{Nucleus$\uparrow$} 
& \multicolumn{3}{c|}{Boundary$\uparrow$} 
& \multicolumn{3}{c|}{Span$\uparrow$} 
& \multirow{2}{*}{Tok/s$\downarrow$} 
& \multirow{2}{*}{RTFx$\uparrow$} \\
\cmidrule(lr){4-6}
\cmidrule(lr){7-9}
\cmidrule(lr){10-12}
& & 
& Prec & Rec & F1 
& Prec & Rec & F1 
& Prec & Rec & F1 
& & \\
\midrule

Audio (raw) & SBS & \texttt{peakdetect} 
& \textbf{90.0} & 92.4 & 91.2 
& 65.1 & 56.8 & 60.6 
& 37.9 & 36.3 & 37.1 
& 3.5 & 684x \\

Audio (raw) & Theta & \texttt{peakdetect}  
& 86.5 & 80.1 & 83.1 
& 52.6 & 41.5 & 46.3 
& 19.7 & 17.0 & 18.2 
& 3.1 & 27x \\

Sylber & Cos. Sim. & Cos. thresh.
& 89.9 & 97.0 & \textbf{93.3} 
& 51.9 & \textbf{80.6} & 63.1 
& \textbf{45.8} & 45.6 & 45.7 
& 3.4 & 36x \\

VG-HuBERT & featSSM & Mincut 
& 78.2 & 89.3 & 83.0 
& 66.1 & 64.8 & 65.0 
& 36.3 & 39.5 & 37.6 
& 3.8 & 6x \\

VG-HuBERT & CLS attn. & CLS thresh. 
& 37.4 & 48.2 & 41.9 
& 40.3 & 44.4 & 42.1 
& 12.0 & 14.7 & 13.2 
& 5.8 & 72x \\
\midrule
Sylber & Cos. Sim. & \texttt{peakdetect}
& 77.4 & 95.6 & 85.5 
& \textbf{67.1} & 73.0 & \textbf{69.9} 
& 43.3 & \textbf{51.6} & \textbf{47.0} 
& 3.9 & 84x \\

VG-HuBERT & Cos. Sim. & \texttt{peakdetect} 
& 65.6 & \textbf{97.3} & 78.3 
& 59.3 & 76.9 & 66.8 
& 35.8 & 51.0 & 42.0 
& 5.1 & 32x \\

VG-HuBERT & CLS attn. & \texttt{peakdetect} 
& 53.7 & 83.7 & 65.3 
& 46.1 & 60.9 & 52.4 
& 17.9 & 25.6 & 21.0 
& 5.5 & 9x \\

VG-HuBERT & featSSM & \texttt{peakdetect} 
& 48.7 & 77.3 & 59.7 
& 40.4 & 52.5 & 45.6 
& 12.7 & 18.0 & 14.9 
& 5.5 & 29x \\

Sylber & featSSM & \texttt{peakdetect} 
& 52.8 & 54.4 & 53.6 
& 46.8 & 40.3 & 43.3 
& 13.6 & 12.7 & 13.1 
& 3.8 & 58x \\

\bottomrule
\end{tabular}
\end{table*}

\section{Experiments}
\subsection{Experimental setup}
We compare two classical envelope-based baselines (SBS, Theta) with three widely used neural configurations (Sylber; VG-HuBERT\textsubscript{\texttt{featSSM}}+MinCut; VG-HuBERT\textsubscript{\texttt{CLS}}+CLS-threshold), and additional configurations 
pairing each feature-based cue with \texttt{peakdetect} via pseudo-envelope export 
(Section~\ref{subsec:segmenters-and-modularity}).
%To enable interoperability with envelope-based segmentation methods, we export SSL-derived time-series traces (e.g., local framewise cosine similarity, CLS-attention trace, or featSSM-derived global coherence traces) as pseudo-envelopes and apply the same \texttt{peakdetect} procedure used for classical envelopes.  For featSSM, the global coherence trace is computed as the per-frame row-wise mean of the SSM, yielding a 1D similarity-to-utterance time series suitable for peak-based segmentation.

%(envelope amplitude, local framewise cosine similarity, CLS-attention trace, or a featSSM-derived global coherence trace). For featSSM, the global coherence trace is computed as the per-frame row-wise mean of the SSM, yielding a 1D similarity-to-utterance time series suitable for peak-based segmentation.

All methods operate on mono 16\,kHz audio normalized by the toolkit utilities and use recommended default hyperparameters where applicable, including auto-calibrated lookahead for \texttt{peakdetect} and an expected syllable duration of 220\,ms for MinCut-based methods. Neural configurations are run sequentially under a single fixed hardware/software setup (Apple M1 Max, PyTorch MPS backend), while classical baselines use CPU parallelization where appropriate; runtime metrics are therefore intended for relative comparisons within this setup.

We evaluate segmentation with a 50\,ms matching tolerance at three granularities: \texttt{nucleus} (peak detection) \cite{xieRobustAcousticbasedSyllable2006,zhangSpeechRhythmGuided2009,yuanRobustSpeakingRate2010,obinSyllOMaticAdaptiveTimefrequency2013}, \texttt{boundary} (onset/offset detection) \cite{obinSyllOMaticAdaptiveTimefrequency2013,villingAutomaticBlindSyllable2004,rasanenPrelinguisticSegmentationSpeech2018,pengSyllableDiscoveryCrossLingual2023,choSDHuBERTSentenceLevelSelfDistillation2024,choSylberSyllabicEmbedding2024}, and \texttt{span} (full interval recovery; both onset and offset must match) \cite{pengWordDiscoveryVisually2022,rasanenPrelinguisticSegmentationSpeech2018}. We adopt the 50\,ms tolerance in line with prior work because syllable nuclei and boundaries are not uniquely defined across theories and annotation practices, so some apparent errors may be genuinely ambiguous.  For interval-based methods, we enable consistent nucleus evaluation by deriving a single nucleus time per predicted interval as the argmax of the configuration’s cue time series within that interval.

For each granularity we compute true positives, insertions, deletions (and substitutions for spans), and report precision, recall, and F1. Evaluation runs in batch over matched WAV--TextGrid pairs using tier-aware parsing.  %, producing long-format outputs (dataset $\times$ configuration $\times$ metric) used for corpus-level aggregation.
In addition to accuracy, we measure token rate (tok/s audio) as a proxy for sequence length, and inverse real-time factor (RTFx) as a proxy for computational cost, both measured during nucleus extraction on the same hardware/software setup.

\subsection{Corpora and Annotations}

Our benchmarks span three recording contexts. Adult read speech includes English TIMIT \cite{garofolojohns.TIMITAcousticPhoneticContinuous1993}, the \texttt{train-clean-100} subset of LibriSpeech \cite{panayotovLibrispeechASRCorpus2015}, and Wikipedia Spanish Speech and Transcripts (WikiSpanish) \cite{hernandezmenaWikipediaSpanishSpeech2021}. Child-directed speech includes two English corpora: Brent \cite{brentRoleExposureIsolated2001,ryttingSegmentingWordsNatural2010,rasanenPrelinguisticSegmentationSpeech2018}, and the Philadelphia Home Corpus (PHC) \cite{swingleyPHC2024}; and the Spanish Ornat corpus from CHILDES \cite{lopezornatAdquisicionLenguaEspanola1994} with hand-annotated phone and word boundaries from \cite{swingleyLexicalLearningMay2018} (Ornat-Swingley). %\cite{lopezornatAdquisicionLenguaEspanola1994,swingleyLexicalLearningMay2018}. 
%Ornat-Swingley includes phone and word tiers but no syllable intervals, so we evaluate nucleus detection only for this corpus.

For English datasets (TIMIT, LibriSpeech, PHC \& Brent), we generate syllable boundaries using a syllabified CMU pronunciation dictionary with rule-based fallback via \texttt{syllabify}\footnote{\url{https://github.com/kylebgorman/syllabify}}. For LibriSpeech, we use word and phone alignments \cite{lugoschLibriSpeechAlignments2019} released by \cite{lugoschSpeechModelPretraining2019} produced with the Montreal Forced Aligner \cite{mcauliffeMontrealForcedAligner2017a}. For WikiSpanish, we use \texttt{faseAlign}\footnote{\url{https://github.com/EricWilbanks/faseAlign}} to obtain syllable-level annotations. The Ornat-Swingley corpus provides phone and word tiers but no syllable 
boundary intervals; since boundary and span evaluation require reference intervals, only 
nucleus detection is evaluated for this corpus.  For Kono, an underdocumented Central Mande language in Sierra Leone, we use fieldwork recordings and orthographic transcriptions from \cite{Hamo2026kono} and manually add word-, phone-, and syllable-level time alignments.\footnote{Kono materials courtesy of Alexander Hamo: \url{https://alexanderhamo.org/kono/index.html}.}

We note that syllable boundaries for English and Spanish are derived algorithmically from forced-aligned phone tiers via pronunciation dictionaries; the resulting syllable-level annotations are intended as a consistent reference for comparative evaluation across methods rather than as an authoritative ground truth on where the indeterminate syllable boundaries lie.

%Table~\ref{tab:data} summarizes the corpora used for evaluation. 

\section{Results and Discussion}

\begin{figure*}[t]
    \centering
    \includegraphics[width=\linewidth]{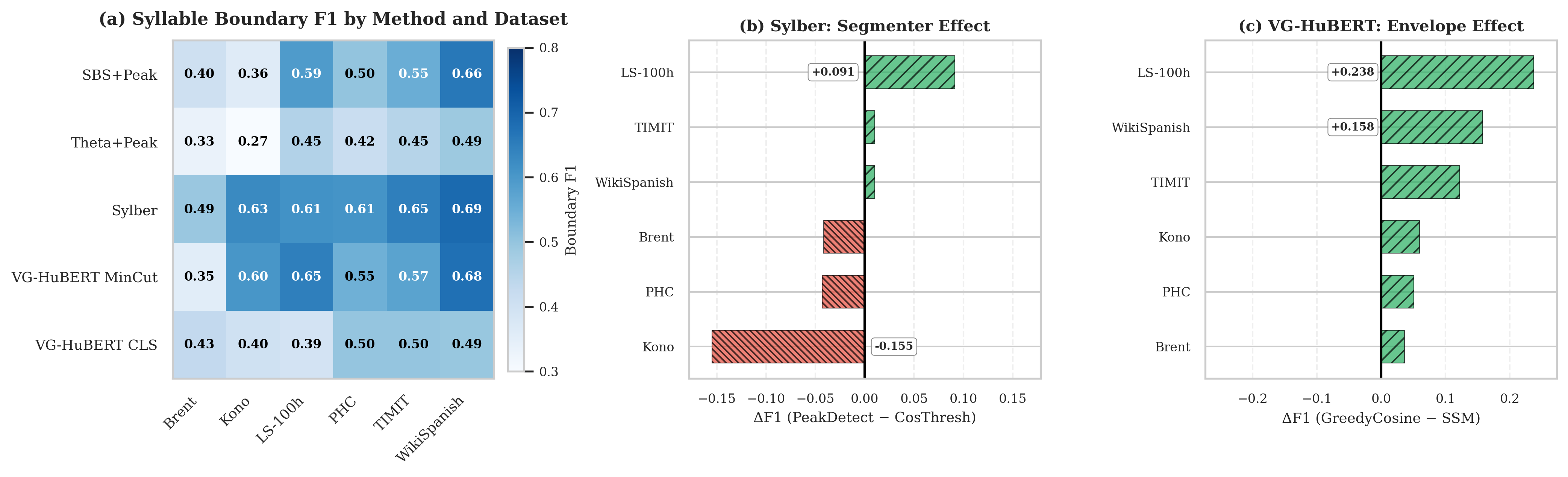}
    \caption{Syllable boundary segmentation across datasets. (a) Boundary F1 by method and dataset for envelope baselines and default published neural configurations (higher is better; Ornat-Swingley excluded due to missing syllable intervals). (b) Sylber component effect: $\Delta$F1 when swapping the segmenter (CosThresh $\rightarrow$ \texttt{peakdetect}) with the cosine-similarity cue held fixed. (c) VG-HuBERT component effect: $\Delta$F1 when swapping the cue/envelope (SSM $\rightarrow$ GreedyCosine) with the \texttt{peakdetect} segmenter held fixed; positive values indicate improvement from the substituted component.}  \label{fig:boundary_segmentation}
\end{figure*}

%\begin{figure}[t]
%  \centering
%  \includegraphics[width=\linewidth]{figures/figure1_boundary_f1_heatmap.png}
%  \caption{Syllable boundary detection performance (F1) by method and dataset (syllable-weighted; higher is better). Methods are the two envelope baselines (SBS+Peak, Theta+Peak) and three default published neural configurations (Sylber, VG-HuBERT MinCut, VG-HuBERT CLS).  The Ornat-Swingley dataset is excluded due to lack of syllable-level annotations.}
%  \label{fig:boundary_f1_heatmap}
%\end{figure}

%\begin{figure}[t]
%  \centering
%  \includegraphics[width=\linewidth]{figures/figure2_method_recombination.png}
%    \caption{Boundary-F1 effects of modular recombination by dataset (syllable-weighted). Top: Sylber, swapping segmenter (CosThresh $\rightarrow$ \texttt{peakdetect}) on the same cosine-similarity cue; $\Delta$F1 = \texttt{peakdetect} $-$ CosThresh. Bottom: VG-HuBERT, swapping envelope/cue (SSM $\rightarrow$ GreedyCosine) with a fixed \texttt{peakdetect} segmenter; $\Delta$F1 = GreedyCosine $-$ SSM. Positive $\Delta$F1 indicates improvement.}  
%    \label{fig:perf_delta}
%\end{figure}

\subsection{Segmentation accuracy}
Table~\ref{tab:full_method_comparison} reports aggregate Precision, Recall, and F1 for nucleus, boundary, and span detection across seven corpora, where metrics 
are weighted by each corpus's syllable count.  Figure~\ref{fig:boundary_segmentation}a (left) shows boundary F1 broken down by dataset for the default baselines/configurations.
Across configurations, nucleus detection is substantially easier than recovering full spans: high nucleus F1 does not necessarily translate into accurate boundary placement and interval recovery.
For example, the SBS envelope with \texttt{peakdetect} attains 91.2 nucleus F1, but drops to 60.6 boundary F1 and 37.1 span F1, indicating that errors compound when moving from local peak detection to full interval reconstruction.

Among the published default configurations, end-to-end syllabifiers provide the strongest overall accuracy.
Sylber with a cosine-similarity envelope and cosine-threshold segmentation yields the best nucleus performance (93.3 F1) and strong span recovery (45.7 F1), while VG-HuBERT with its feature self-similarity matrix (featSSM) and MinCut achieves the best default boundary performance (65.0 F1).
The VG-HuBERT CLS-attention configuration performs poorly in our setting (e.g., 41.9 nucleus F1); given potential sensitivity to implementation and tuning choices, we report it as an observed baseline behavior rather than attributing a specific cause.

Segmentation quality depends on both (i) the representation/envelope and (ii) the segmentation algorithm, and (iii) the interactions between the two.
Mixing and matching components reveals improvements (Figure~\ref{fig:boundary_segmentation}b \& \ref{fig:boundary_segmentation}c) that are not visible when methods are evaluated only as packaged pipelines.
Applying \texttt{peakdetect} to Sylber's cosine-similarity envelope improves boundary detection from 63.1 to 69.9 F1 (Figure~\ref{fig:boundary_segmentation}b, middle) and span recovery from 45.7 to 47.0 F1, yielding the best boundary and span scores in the table.
For VG-HuBERT, swapping the featSSM-derived envelope for local cosine similarity over layer-8 features changes the boundary/span operating point under the same \texttt{peakdetect} segmenter, improving span recovery while keeping boundary performance competitive (Figure~\ref{fig:boundary_segmentation}c, right). 
Overall, these results support a modular toolkit: practical gains can come from recombining representations, envelopes, and segmenters rather than treating each syllabifier as a fixed tokenizer.

\subsection{Token rate and computational cost}
Table~\ref{tab:full_method_comparison} also reports token rate (tok/s audio) and inverse real-time factor (RTFx), where RTFx is computed as total audio duration divided by wall-clock time (higher is faster). Because RTFx is hardware- and implementation-dependent, we report it primarily to illustrate relative throughput differences between configurations measured under the same conditions (Apple M1 Max), i.e., the scaling trend within Table~\ref{tab:full_method_comparison}.
All approaches produce only a few syllabic tokens per second of audio (3.1--5.8 tok/s), representing a large reduction relative to typical 50--100~Hz frame-level features and making syllabic tokenization attractive for long-context spoken language modeling.

Methods nevertheless differ sharply in throughput.
Envelope-based baselines are consistently the most computationally efficient, while neural representations incur higher cost but can yield improved boundary placement and span recovery.
Within the neural family, we observe an accuracy--throughput trade-off: configurations that improve boundaries and spans generally reduce throughput relative to simple peak-based envelopes.
Conversely, if an application primarily requires nuclei (rather than full spans), fast envelope-based configurations provide a strong operating point with only a small reduction in nucleus F1 relative to the best-performing neural configuration.
Overall, \texttt{findsylls} enables selecting operating points along the accuracy--throughput curve depending on whether the application prioritizes streaming speed, boundary fidelity, or full span recovery.

\subsection{Limitations \& Future Work}
Some configurations are sensitive to implementation and tuning choices; while we follow published descriptions where possible, we do not claim exact reproduction for every prior system.
RTFx is measured on a single hardware/software platform and should be interpreted comparatively.
In addition, the current release does not yet include all methods surveyed in related work; expanding coverage and including preset configurations is a primary goal of future iterations. 
%Finally, syllable boundaries are sometimes genuinely ambiguous, so a subset of measured boundary errors may reflect inconsequential placement differences rather than clear segmentation mistakes. We will manually audit a sample of errors and develop refined evaluation criteria that distinguish plausible alternatives from true failures.
Finally, the current evaluation does not yet implement tolerance regions around linguistically indeterminate boundary placements (e.g., word boundaries, consonantal clusters), so a subset of measured errors likely reflects genuine annotation ambiguity rather than clear segmentation failures; future versions will incorporate refined evaluation criteria and error audits to more accurately assess each configuration's performance.

\section{Conclusion}
%We introduced \texttt{findsylls} (v1), a modular toolkit for syllable nucleus, boundary, and span segmentation, and benchmarked ten configurations across seven corpora using syllable-weighted metrics.
%We introduced \texttt{findsylls} (v1), a language-agnostic, modular toolkit for syllable-level speech tokenization and embedding, with a unified interface for nucleus, boundary, and span segmentation and multi-granular evaluation.
%We demonstrated its capabilities by benchmarking ten classical and representation-driven configurations across seven corpora under matched metrics. %, reporting both segmentation quality and token-rate/runtime characteristics.
%Our results highlight a clear speed--accuracy trade-off: simple envelope baselines can deliver strong nucleus accuracy at high RTFx, while improved boundaries and spans benefit from stronger learned representations and segmentation choices.
%Recombining components yields measurable gains over published defaults, supporting the view that syllabic tokenization is a design space rather than a single fixed method.
%\texttt{findsylls} is designed as an extensible framework for reproducible evaluation and rapid method composition.
%Future work will expand method coverage and strengthen reproducibility (e.g., hyperparameter presets) across additional languages and domains.
%\texttt{findsylls} is intended as shared infrastructure for syllable-level tokenization and embedding, enabling reproducible evaluation and controlled method composition across languages and domains.

In this paper, we demonstrate the utility of \texttt{findsylls} (v1) as a syllable segmentation and processing toolkit by benchmarking ten syllabification pipelines and modular recombinations thereof across seven corpora and three languages under matched conditions. Ongoing work expands methodological coverage and functionality toward consolidating decades of research on signal processing and neural methods of unsupervised syllable detection into a single interface with a common vocabulary. \texttt{findsylls}'s ultimate goal is to empower speech researchers with the full design space of syllable processing, and to push the frontier of speech and language model research toward more efficient representations of speech.

\section{Acknowledgements}
I thank my advisors Charles Yang, Daniel Swingley, Lyle Ungar, and Mark Liberman for their guidance throughout this work. In particular, I am grateful to Charles Yang and Daniel Swingley for their insights into the developmental role of syllables in infant language acquisition; to Daniel Swingley for sharing his hand-annotated PHC and Ornat-Swingley corpora, which enabled the detailed comparisons presented here; and to Mark Liberman for sharing his earlier work on signal-based syllable detection, and for being the first user of \texttt{findsylls} after myself.

I am also grateful to the University of Pennsylvania Linguistics community, whose questions, comments, and discussions led to the formalization of the present work.

Lastly, I thank the Interspeech 2026 reviewers, whose constructive and detailed feedback greatly improved the present paper and the \texttt{findsylls} development roadmap.

\section{Generative AI Use Disclosure}

Generative AI was used for copyediting and polishing the manuscript. All technical content, experimental design, results interpretation, and conclusions are my own work. I assume full responsibility for the accuracy and integrity of the paper and consent to its submission.

\bibliographystyle{IEEEtran}
\bibliography{references,mybib}

\end{document}